\documentclass[11pt]{article}

\usepackage[preprint]{acl}
\usepackage{amsmath}
\usepackage{multirow}
\usepackage{times}
\usepackage{latexsym}
\usepackage{booktabs}
\usepackage{xspace}
\usepackage{hyperref}
\usepackage{graphicx}
\usepackage{subcaption}
\usepackage{tabularx}
\usepackage{float}
\usepackage[T1]{fontenc}
\usepackage{listings}
\usepackage{xcolor}

\lstdefinestyle{jsonstyle}{
    basicstyle=\ttfamily\footnotesize,
    frame=none,
    breaklines=true,
    columns=fullflexible,
    keepspaces=true,
    showstringspaces=false,
    xleftmargin=0.5em,
    xrightmargin=0.5em
}
\usepackage[utf8]{inputenc}

\usepackage{microtype}

\usepackage{inconsolata}

\usepackage{graphicx}

\newcommand{\tool}{\texttt{CoDA}\xspace}

\title{\textsc{CoDA}: A Human-LM Collaborative \emph{Co}debook \emph{D}evelopment \emph{A}ssistant for Domain-Specific Text Annotation}

\author{
 \textbf{Jaspreet Ranjit\textsuperscript{1}},
 \textbf{Swaminathan Chellappa\textsuperscript{1}},
 \textbf{Augusto Rivas Costante\textsuperscript{1}},
\\
 \textbf{Chandan Manjunath\textsuperscript{1}},
 \textbf{Laura Onasch-Vera\textsuperscript{2}},
 \textbf{Hailey Winetrobe Nadel\textsuperscript{3}},
 \\
 \textbf{Eric Rice\textsuperscript{3}},
 \textbf{Swabha Swayamdipta\textsuperscript{1}}
\\
\\
 \textsuperscript{1}Thomas Lord Dept. of Computer Science, University of Southern California,\\
 \textsuperscript{2}My Friend's Place, Los Angeles, California\\
 \textsuperscript{3}School of Social Work, University of North Carolina
\\
 \small{
   \textbf{Correspondence:} \href{mailto:jranjit@usc.edu}{jranjit@usc.edu}
 }
}

\begin{document}
\pagestyle{empty}
\maketitle
\begin{abstract}

We present \tool, an open-source human-LM collaborative \emph{co}debook \emph{d}evelopment \emph{a}ssistant that helps domain experts develop annotation guidelines for annotating new variables in unstructured text with language models.
Existing language model annotation systems primarily focus on downstream classification performance, while assuming the availability of high-quality annotation guidelines, or codebooks. 
In practice, developing these codebooks is time-consuming and challenging, particularly for concepts that require expert knowledge.
\tool addresses this gap by collaborating with domain experts to iteratively develop codebooks through targeted review of incorrect language model predictions and incorporating expert feedback into the evolving codebook.
We demonstrate \tool on two real-world case studies: extracting health indicators from homelessness case management notes and classifying sentiment of wildlife trafficking posts on social media. 
Across both domains, \tool produces codebooks that achieve downstream classification performance comparable to expert codebooks while reducing codebook development time from several weeks to two hours and requiring 5$\times$ fewer manually annotated examples.
Applying \tool to generate annotations at scale, we identified 2,410 case notes that contained health discussions or referrals and 10,196 social media posts expressing negative sentiment that may warrant investigation for potential wildlife trafficking.
By accelerating codebook development, \tool helps practitioners use language models to extract domain-specific insights from unstructured text at scale.

\end{abstract}
\section{Introduction}
\label{sec:intro}
\begin{figure*}[h!]
  \includegraphics[width=\textwidth]{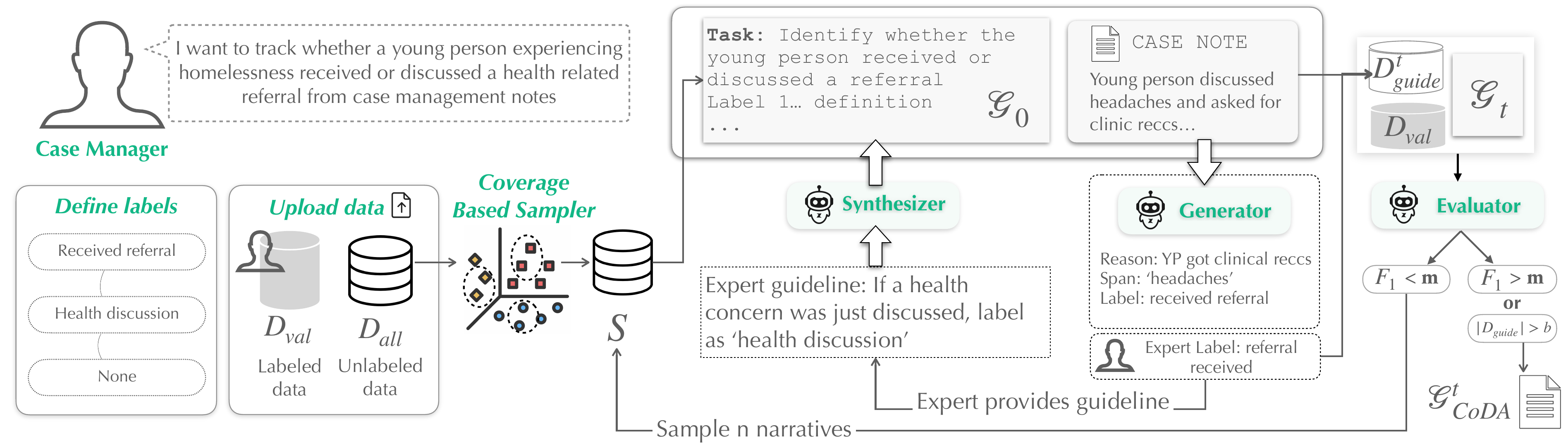}
  \caption{\tool is a human-in-the-loop system for developing annotation guidelines (codebooks) and scaling domain-specific text annotation. Given a small labeled validation set ($\mathcal{D}_{\text{val}}$) and an unlabeled dataset $\mathcal{D}_{\text{all}}$, the \emph{Sampler} selects representative samples from $\mathcal{D}_{\text{all}}$ using coverage based sampling. 
  The \emph{Generator} uses a language model ($\mathcal{M}_{\text{\tool}}$) to predict a label, supporting span, and reasoning. Experts provide feedback on incorrect predictions. The \emph{Synthesizer} uses $\mathcal{M}_{\text{\tool}}$ to incorporate expert feedback into an evolving codebook ($\mathcal{G}^{t}_{\text{\tool}}$), and the \emph{Evaluator} prompts $\mathcal{M}_{\text{\tool}}$ with $\mathcal{G}^{t}_{\text{\tool}}$ to generate labels and measure Macro F$_1$ on $\mathcal{D}_{\text{val}}$ after each iteration. Once the Macro F$_1$ on $\mathcal{D}_{\text{val}}$ reaches an expert-defined threshold $m$, or the expert has exhausted the budget $b$, $\mathcal{G}^{t}_{\text{\tool}}$ is saved at iteration $t$ for large scale annotation. For our evaluated case studies, \tool reduces codebook development from several weeks to approximately two hours while requiring 5$\times$ fewer expert-reviewed examples.}

  \label{fig:alg}
  \vspace{-0.5cm}
\end{figure*}


Human-centered occupations rely on unstructured text data (e.g., clinical notes, social media posts, and case notes) to track outcomes for the populations they serve. 
For example, healthcare workers track health indicators from electronic medical records to inform evidence-based care~\cite{guevara2024large}, whereas social workers track theory of change indicators in case notes to inform interventions for youth experiencing homelessness~\cite{nelson2024designing}. 
Extracting these insights is challenging due to the scale of the data, the need for domain expertise to infer implicit signals, and the private and sensitive nature of the text.

Language models (LMs) provide a promising avenue for scaling the annotation of text data.
Existing approaches formulate annotation as a classification task, prompting LMs with a codebook, or set of annotation guidelines, that defines a target concept and specifies how text should be assigned to a set of labels~\citep{glaser2017discovery}. 
Prior work has largely focused on improving the downstream classification performance within this setup. However, this workflow hinges on a tacit assumption: that a reliable codebook already exists and faithfully captures the concept the domain expert wishes to measure~\citep{schroeder2025just}. 

Developing codebooks in domain-specific settings is particularly challenging given that the target concepts are often nuanced and take on canonical meanings that are not widely understood~\citep{halterman2026protest, kim2026refining, chen2026effects}.
For example, in homelessness, a \emph{health referral} can include not only direct referrals to healthcare providers, but also assistance such as arranging transportation to a medical appointment or scheduling care.
When LMs use codebooks to annotate such domain-specific concepts, errors can arise from either (1) models misapplying instructions (scoring errors) or (2) codebooks that fail to capture the expert's intended concept (conceptualization errors)~\citep{wallach2025position}.
While better models may reduce scoring errors, conceptualization errors require refining the codebook itself.

\begin{figure*}[h!]
  \includegraphics[width=\textwidth]{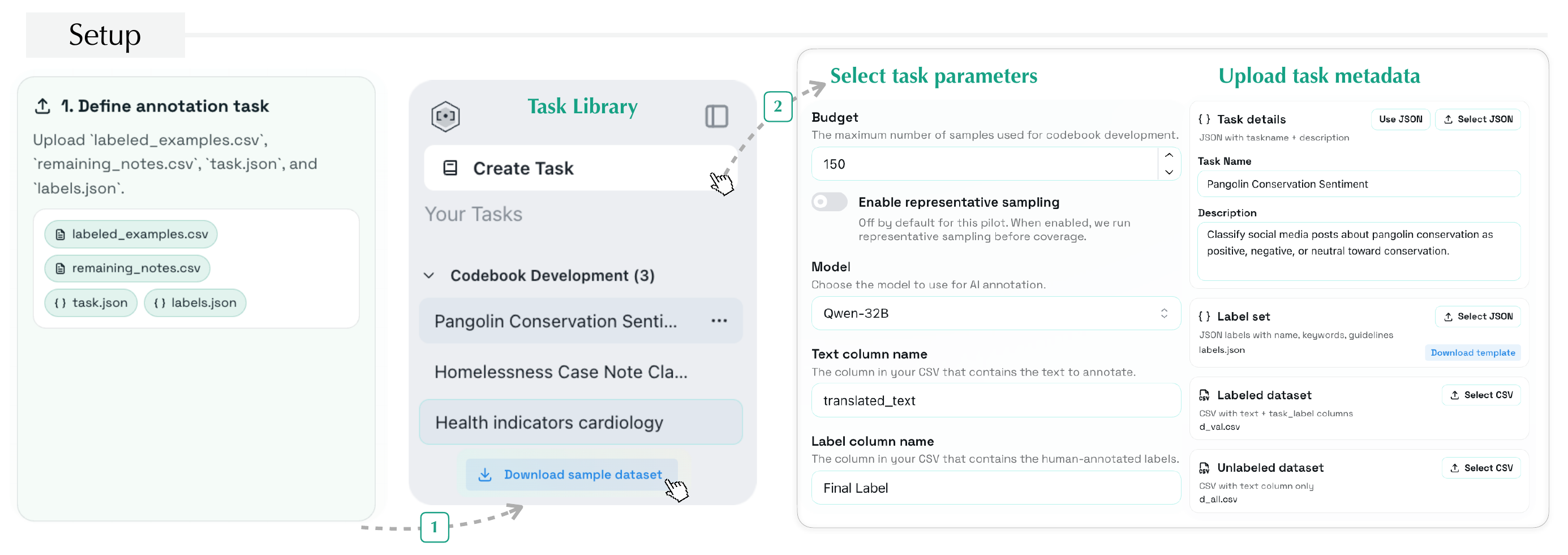}
  \caption{Setting up an annotation task in \tool. (1) Users create a new task from the task library, where they can access existing tasks or initialize a new task for codebook development. (2) Users configure the annotation budget, sampling strategy, and language model, and upload the task metadata and datasets. Task metadata specifies the annotation task and label set; the labeled dataset serves as the validation set ($\mathcal{D}_{\text{val}}$) for evaluating codebooks during development, while the unlabeled corpus ($\mathcal{D}_{\text{all}}$) contains the examples to be annotated at scale.}

  \label{fig:userflow_setup}
  \vspace{-0.5cm}
\end{figure*}

To address these challenges, we introduce \tool, an open-source human-LM collaborative system for iterative codebook development and large-scale text annotation. 
As illustrated in \autoref{fig:userflow_setup} and \autoref{fig:userflow_main}, the user workflow consists of creating an annotation task, iteratively refining annotation guidelines through human-LM collaboration, evaluating the evolving codebook on a held-out validation set, and scaling annotation to the remaining unlabeled corpus. 
Our system consists of four modules: the \textit{Sampler}, \textit{Generator}, \textit{Synthesizer}, and \textit{Evaluator}, which identify diverse examples, generate model predictions, synthesize expert feedback into an evolving codebook, and monitor the Macro F$_1$ on a held out validation set throughout development (\autoref{fig:alg}).
We release \tool\ as both a web-based application and a locally deployable package to support annotation workflows involving sensitive data. 
Across our two real-world case studies in youth homelessness and wildlife conservation, \tool produces codebooks that achieve downstream classification performance comparable to expert codebooks while significantly reducing expert annotation effort. Specifically, in the homelessness case study, compared with the fully manual codebook development process, \tool reduced development time from several weeks to approximately two hours and required 5$\times$ fewer manually annotated examples.
Finally, we prompt LMs with \tool codebooks to scale annotations and identify 1,474 homelessness case notes containing health discussions, 936 containing health referrals, and 10,196 social media posts with negative sentiment.
These case studies demonstrate how \tool enables practitioners to transform large collections of unstructured text into structured insights that support evidence-based decision making in high-impact domains.

\section{Design and Implementation}
\label{sec:design}

\tool has three stages: (1) task definition, (2) human-in-the-loop codebook development and (3) scaling annotations.
In this section, we elaborate on each of the stages, include details on the installation/setup, and describe our implementation, including the software stack and deployment modes. Additionally, we conducted codesigns with three community organizations to inform the design of \tool. We include our findings in Appendix \textsection \ref{app:codesign}. 

\subsection{Installation and Setup}
\tool is available as a web-based application hosted on AWS\footnote{\url{https://coda.44.218.57.167.sslip.io}}. The implementation is open-source and publicly available on GitHub\footnote{\url{https://github.com/jr4fs/CoDA/}} which includes detailed installation instructions for deploying \tool on a user's preferred machine and a video walkthrough \footnote{\url{https://youtu.be/XdJ3DEFMawE}}. 
Given our focus on societally impactful applications, which often involve sensitive or private data, \tool can also be deployed locally as a desktop application using language models served through Ollama or hosted entirely on infrastructure managed by the user.
\tool is designed for both researchers and community practitioners, and therefore, we provide multiple onboarding modes tailored to different use cases. First, we provide a static \textit{Demo} mode that walks users through the human-in-the-loop codebook development interface using a lightweight pre-populated task. 
Second, we provide a fully interactive end-to-end tutorial that guides users through creating a task, iteratively refining a codebook, and scaling annotations using a toy dataset\footnote{\url{https://github.com/jr4fs/CoDA/tree/main/sample_dataset}}. 

\subsection{Implementation}
\tool is implemented as a full-stack web application with a React/TypeScript frontend, a Node.js/Express backend for application logic, and a Python/FastAPI backend. The backend performs embedding-based sampling, language model inference, and implements the human-in-the-loop codebook development algorithm (\autoref{fig:alg}) using PyTorch, Sentence Transformers, FAISS, and Ollama while MongoDB stores annotations, task metadata, user accounts, and evaluation metrics. See Appendix \textsection \ref{app:prompts_software} for package specifications. 

To support different deployment settings, \tool can be deployed locally using open-source language models with Ollama, or configured to use hosted inference providers through a common model interface as documented in our GitHub repository. This modular architecture enables deployment across local machines, academic computing infrastructure, or AWS cloud environments without modifying the user-facing workflow.

\subsection{Task Definition}
Each task in \tool is formulated as a multi-class classification task and corresponds to a single codebook defining a target variable. Each variable is associated with a set of labels (e.g., \emph{Variable}: characterization of health referrals for youth experiencing homelessness; \emph{Labels}: \texttt{received referral}, \texttt{health discussion}, \texttt{none}).

Users can create an account and manage multiple tasks through their task library as illustrated in \autoref{fig:userflow_setup}. To create a new task, the user uploads a \texttt{task.json} file containing the task description and a \texttt{labels.json} file specifying the label set. Optionally, each label may also include a description, keywords, and seed annotation guidelines to initialize the codebook.

The user then uploads two datasets: (1) a human-annotated validation set ($\mathcal{D}_{val}$), which is held out for evaluation, and (2) an unlabeled corpus ($\mathcal{D}_{all}$) from which samples are iteratively selected for annotation and codebook refinement. Finally, the user specifies several task-level parameters, including (1) the annotation budget ($b$), which determines the maximum number of examples allocated to human-in-the-loop codebook development; (2) the language model used for annotation and codebook development; and (3) metadata describing the columns in the uploaded datasets. Once initialized, the task is added to the user's task library, allowing progress to be saved and resumed across sessions. 

\begin{figure*}[h!]
  \includegraphics[width=\textwidth]{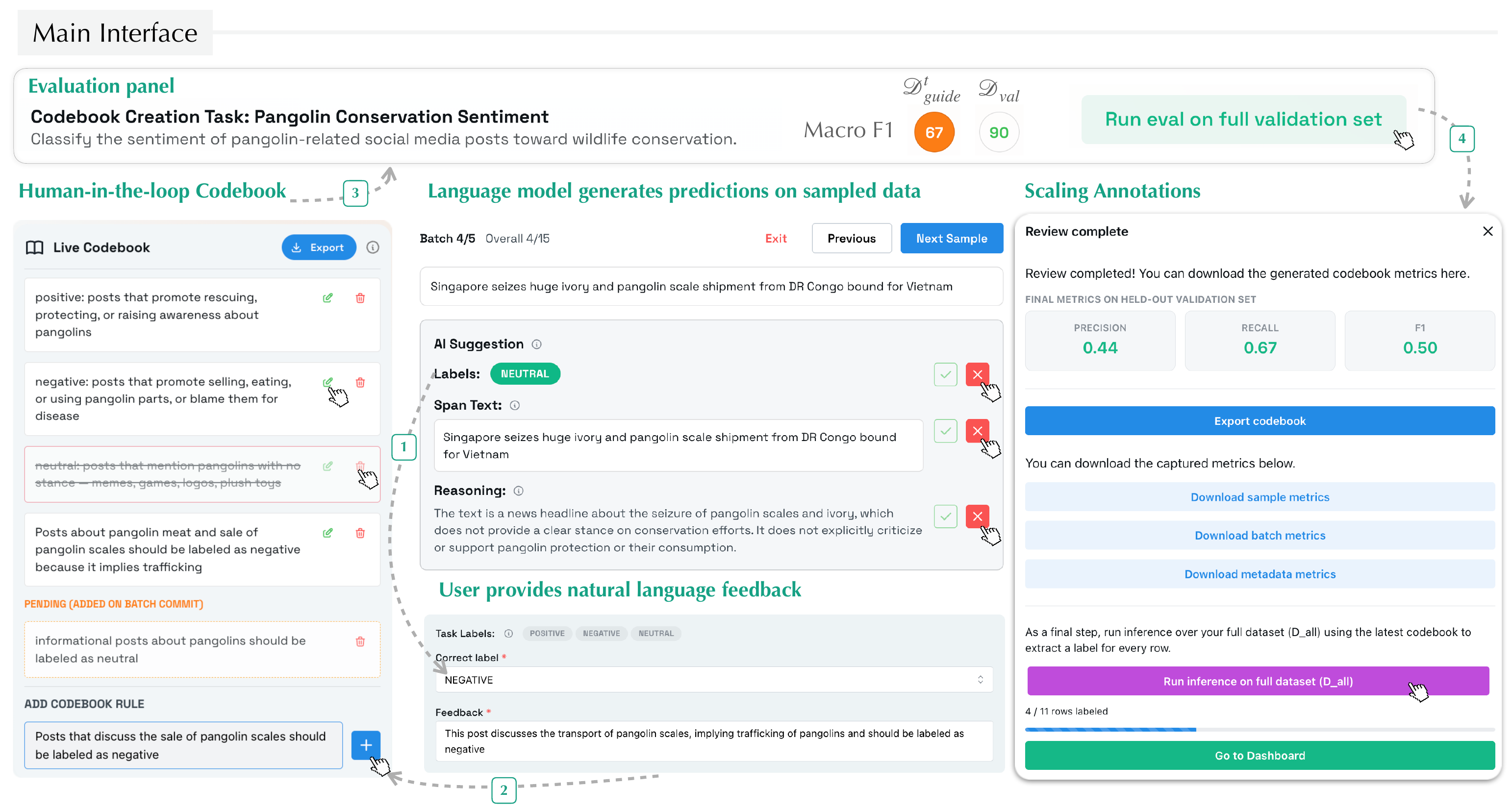}
  \caption{\tool's main codebook development interface. (1) During each iteration, the language model generates a predicted label, supporting text span, and chain-of-thought reasoning for a sampled example, which the user reviews. (2) For incorrect predictions, the user provides the corrected label and natural-language feedback, which \tool synthesizes into the evolving codebook shown in the left panel. (3) Throughout codebook development, users can inspect and edit the evolving codebook and monitor its Macro F$_1$ performance on the held-out validation set $\mathcal{D}_{\text{val}}$. (4) After completing codebook development, users can export the codebook and evaluation metrics and use the resulting codebook to annotate the full unlabeled corpus $\mathcal{D}_{\text{all}}$ at scale.}

  \label{fig:userflow_main}
  \vspace{-0.5cm}
\end{figure*}

\subsection{Human-in-the-loop Codebook Development}
The human-in-the-loop codebook development builds on the algorithm introduced by \citet{ranjit2025interventions}. As illustrated in \autoref{fig:alg}, \tool consists of four modules: a \emph{sampler}, \emph{generator}, \emph{synthesizer}, and \emph{evaluator}. \autoref{fig:userflow_main} shows the corresponding user interface.
Because the unlabeled corpus $\mathcal{D}_{all}$ can be large and the prevalence of target variables is often low~\citep{ranjit2025interventions}, \tool first retrieves examples relevant to the user-defined keywords using FAISS similarity search \citep{douze2024faiss}. For each label, the top 10\% most similar examples are selected to construct a candidate pool from $\mathcal{D}_{all}$.
At iteration $t$, the \emph{sampler} selects a batch of $n$ examples from the candidate pool for expert review, which are added to the accumulating set: $\mathcal{D}^{t}_{\text{guide}}$. We use coverage-based sampling \citep{gupta-etal-2023-coverage} to prioritize examples that are dissimilar to those reviewed in previous iterations, encouraging codebook refinement on diverse and representative cases.

For each sampled example, the \emph{generator} module prompts a language model ($\mathcal{M_{\text{\text{\tool}}}}$) with the current codebook $\mathcal{G}^{t}_{\text{\tool}}$ along with the task and label metadata provided by the user to generate (1) a predicted label, (2) a supporting text span, and (3) a chain-of-thought reasoning. 
The expert then verifies each component, marking it as correct or incorrect. 
When a prediction is incorrect, the user provides natural language feedback correcting the error. The \emph{synthesizer} module uses $\mathcal{M_{\text{\tool}}}$ to incorporate this feedback into the codebook by revising existing guidelines and consolidating redundant or overlapping rules.

At the end of each iteration, the \emph{evaluator} prompts $\mathcal{M_{\text{\tool}}}$ with the codebook at iteration $t$ ($\mathcal{G}^{t}_{\text{\tool}}$) to generate labels on $\mathcal{D}_{val}$ and reports Macro F$_1$ on both $\mathcal{D}_{val}$ and $\mathcal{D}^{t}_{\text{guide}}$. The refinement loop continues until Macro F$_1$ has reached a user-defined threshold $m$, or the user has exhausted the budget $b$. At any point, the users may also manually edit, delete, or add guidelines to the codebook, with changes taking effect immediately. See Appendix \textsection \ref{app:prompts_software} for \tool prompts.

\subsection{Scaling Annotations}
After completing codebook development, users can generate labels for the full unlabeled corpus $\mathcal{D}_{all}$ by prompting a language model with the final codebook $\mathcal{G}^{t}_{\text{\tool}}$. The final codebook is selected from the iteration at which either the user-defined Macro F$1$ threshold $m$ on $\mathcal{D}_{val}$ is reached or the annotation budget $b$ is exhausted.
The system also allows users to export the final codebook along with evaluation metrics (e.g., precision, recall, and Macro F$_1$ on $\mathcal{D}_{val}$ and $\mathcal{D}^{t}_{\text{guide}}$), user feedback, codebook snapshots across iterations, and timing statistics for each iteration.

\section{Validation and Usage}
\label{sec:validation}
We evaluate \tool on two real world case studies in youth homelessness and wildlife trafficking. 
In our first case study, we use \tool to develop a codebook for identifying whether youth experiencing homelessness (YEH) received physical health-related referrals from a corpus of 6,725 case management notes. 
In our second case study, we use \tool to develop a codebook for classifying the sentiment of social media posts about pangolins as a proxy for identifying covert instances of wildlife trafficking.

For each case study, we first ask experts to manually develop a codebook ($\mathcal{G}_{\text{expert}}$) and use this codebook to annotate a held out validation set $\mathcal{D}_{\text{val}}$. 
Then, we ask the experts to use \tool to develop a codebook for the same task ($\mathcal{G}^{t}_{\text{\tool}}$) on a disjoint set of examples from those used to develop $\mathcal{G}_{\text{expert}}$.
In both case studies, experts annotate 150 examples for codebook development and are shown 5 examples per iteration.
For evaluation, we select the codebook from iteration $t$ that achieves the highest Macro F$_1$ on $\mathcal{D}_{\text{val}}$. We use Qwen1.5-32B-Chat with Q4\_0 GGUF quantization as $\mathcal{M_{\text{\tool}}}$ in both case studies.
As a zero-shot baseline, we also evaluate a codebook with only the task definition, and the label descriptions as provided by the expert ($\mathcal{G}^{0}$). 
Using each codebook in the prompt, we compare the Macro F$_1$ on $\mathcal{D}_{\text{val}}$ across three models.

\subsection{Results}
We report Macro F$_1$ on $\mathcal{D}_{\text{val}}$ across Llama-3-70B, Qwen2.5-32B, and Qwen2.5-14B by prompting the model with either $\mathcal{G}_{\text{expert}}$, $\mathcal{G}^{t}_{\text{\tool}}$, or 
$\mathcal{G}^{0}$ (\autoref{tab:main_results}).
Overall, \tool produces codebooks that achieve performance comparable to expert-authored codebooks while consistently outperforming the zero-shot baseline and reducing codebook development from several weeks to approximately two hours using 5$\times$ fewer annotated examples.
Although all codebooks were developed using Qwen1.5-32B-Chat, they serve as useful guidelines for Llama-3-70B, Qwen2.5-32B, and Qwen2.5-14B, suggesting that codebooks may generalize across language models. 
Importantly, \tool substantially reduces expert effort. In the homelessness case study, experts developed a codebook that performed on par with the manually developed codebook in approximately two hours from just 115 annotated examples, whereas the manual codebook required several weeks of expert effort and 593 annotated examples.
Qualitatively, \tool codebooks preserve the intent of the expert guidelines while introducing more explicit decision boundaries for ambiguous cases (see \autoref{tab:mfp_codebook} in Appendix \textsection \ref{app:casestudy}).


\begin{table*}[t]
\centering
\small
\begin{tabular}{lccc|ccc}
\toprule
& \multicolumn{3}{c|}{\textbf{My Friend's Place}} & \multicolumn{3}{c}{\textbf{Wildlife Trafficking}} \\
\cmidrule(lr){2-4} \cmidrule(l){5-7}
\textbf{Model} &
$\mathcal{G}^{0}$ &
$\mathcal{G}^{23}_{\text{\tool}}$ &
$\mathcal{G}_{\text{expert}}$ &
$\mathcal{G}^{0}$ &
$\mathcal{G}^{20}_{\text{\tool}}$ &
$\mathcal{G}_{\text{expert}}$ \\
\midrule
Llama-3-70B & 0.236 & \textbf{0.418} & 0.334 &
0.388 & \textbf{0.556} & 0.547 \\

Qwen2.5-32B & 0.343 & \textbf{0.493} & 0.446 &
0.398 & \textbf{0.552} & 0.529 \\

Qwen2.5-14B & 0.200 & \textbf{0.410} & 0.137 &
0.429 & \textbf{0.560} & 0.519 \\




\bottomrule
\end{tabular}
\caption{Macro F$_1$ of a baseline codebook ($\mathcal{G}^{0}$), the best synthesized codebook developed with \tool\ ($\mathcal{G}^{t}_{\text{\tool}}$), and the expert codebook ($\mathcal{G}_{\text{expert}}$) across both case studies. All \tool codebooks were developed using Qwen1.5-32B. For each case study, we selected the best-performing \tool\ codebook on $D_{\text{val}}$ (iteration $t=23$ for My Friend's Place and $t=20$ for Wildlife Trafficking) and used it to prompt each of the three language models on the held-out validation set. Across all evaluated models, \tool codebooks consistently outperform the simple baseline codebook while achieving performance comparable to, and in several cases exceeding, expert codebooks despite requiring substantially less expert time and annotation effort.}
\label{tab:main_results}
\end{table*}

\subsubsection{Youth Homelessness Case Management}
We collaborated with My Friend's Place, a community organization in Los Angeles serving youth experiencing homelessness (YEH).
After each client interaction, case managers document free-text case notes containing rich context about referrals and client needs. 
Although these notes can inform data-driven interventions, they remain largely unused because extracting structured information from notes requires extensive manual effort.

We use \tool to develop a codebook for identifying whether a youth received a physical health related referral, with three labels: \textit{received referral}, \textit{health discussion}, and \textit{none}. 
We first asked a homelessness data and impact manager and a youth homelessness researcher\footnote{Co-authors of this paper} to independently develop an expert codebook~\citep{glaser2017discovery}.
Over several weeks, the experts iteratively refined the codebook while independently annotating 593 examples, reaching a Krippendorff's $\alpha$ of 0.8~\citep{Krippendorff2011ComputingKA}. We reuse these expert-annotated examples as the held-out validation set ($|\mathcal{D}_{\text{val}}|=593$) for evaluating \tool.
A homelessness data and impact manager then used \tool to develop a codebook for the same task, with an annotation budget $b$ of 150 examples. For the purposes of validation, we retrospectively selected the codebook checkpoint with the highest Macro F$_1$ on $\mathcal{D}_{\text{val}}$, which occurred at the 23rd iteration after the expert reviewed 115 examples.
As shown in \autoref{tab:main_results}, Qwen2.5-32B achieved a Macro F$_1$ of 0.493 when prompted with $\mathcal{G}^{23}_{\text{\tool}}$, and performed comparably when prompted with $\mathcal{G}_{\text{expert}}$. 
This trend also holds across Llama-3-70B and Qwen2.5-14B. 
Finally, we prompted Qwen2.5-32B with $\mathcal{G}^{23}_{\text{\tool}}$ to classify all 6,725 case management notes, and identified 1,474 notes containing health discussions and 936 in which a youth received a physical health referral. 

We compare the expert effort required during the codebook development stage for \tool and the fully manual process. 
In the fully manual process, experts iteratively annotated 593 examples to develop the codebook over several weeks. 
With \tool, the expert instead reviewed model predictions and provided feedback over successive iterations; the best-performing codebook was obtained after reviewing 115 examples, requiring approximately two hours. 
Thus, \tool used approximately 5$\times$ fewer expert-reviewed examples during codebook development. This comparison does not include the cost of annotating $\mathcal{D}_{\text{val}}$, which is used only for evaluation.

\subsubsection{Wildlife Trafficking}
In our second case study, we evaluate \tool on developing a codebook for classifying the sentiment (positive, negative, or neutral) of social media posts about pangolins, one of the world's most trafficked mammals~\citep{murray2023pangolins}. Detecting online wildlife trafficking is challenging because traffickers frequently use coded language, species nicknames, and emojis to conceal illicit activity. Sentiment can serve as a proxy for identifying trafficking-related content; for example, a post such as \textit{"yum, pangolin soup"} expresses positive sentiment but indicates harmful consumption of a protected species.

We collaborated with an AI for wildlife conservation researcher to first develop an expert codebook and annotate a held out validation set ($|\mathcal{D}_{\text{val}}|=585$). 
One of the experts then used \tool to develop a codebook for the same task; at the 20th iteration, the codebook achieved the highest Macro F$_1$ on $\mathcal{D}_{\text{val}}$.
$\mathcal{G}^{20}_{\text{\tool}}$ was developed in approximately one hour after reviewing 100 examples. 
As shown in \autoref{tab:main_results}, Llama-3-70B achieved a Macro F$_1$ of 0.556 when prompted with $\mathcal{G}^{20}_{\text{\tool}}$, while outperforming $\mathcal{G}_{\text{0}}$ and $\mathcal{G}_{\text{expert}}$. 
We prompted Llama-3-70B with $\mathcal{G}^{20}_{\text{\tool}}$ to classify 237K social media posts, and identified 10,196 negative sentiment posts indicative of potential trafficking. 

\section{Related Work}
\label{sec:relatedwork}

Recent advances in language models have enabled the automated classification and annotation of large-scale unstructured text data~\citep{tai2024examination,kirsten2025assistance,gilardi2023chatgpt,tornberg2023chatgpt}.
Converting unstructured text into structured insights is particularly valuable in human-centered domains, where critical information is often documented as free-text notes rather than structured records.
However, these applications frequently require identifying nuanced, canonical concepts that require precise definitions to ensure reliable annotation~\citep{ranjit2024oath, ranjit2025interventions, mendelsohn-etal-2021-modeling, mendelsohn-etal-2023-dogwhistles}. 
While language models show strong promise for scaling annotation~\citep{10.1145/3613904.3641960}, their performance depends heavily on the availability of comprehensive codebooks that precisely define each concept and provide detailed annotation instructions~\citep{barany2024chatgpt, xiao2023supporting, perkins2024use}. 

Developing these codebooks is a labor-intensive process, typically requiring multiple rounds of qualitative coding, expert discussion, and iterative refinement. Moreover, even carefully developed human annotation guidelines often transfer poorly to language models because they are not tailored to the particular ways in which language models fail in domain specific contexts~\citep{pangakis2023automated, xie2024automating, halterman2026codebook, rytting2023towards}. 

Prior work has examined the effectiveness of language models for qualitative coding~\citep{dunivin2024scalable, kim-etal-2024-meganno}, but has not thoroughly examined the development of the codebook guidelines themselves with language models~\citep{dai2023llm, chew2023llm}. The nearest neighbor to our work is the framework introduced by \citet{xiong2025co}, which focuses on surfacing edge cases and reaching thematic saturation. Our underlying algorithm instead combines coverage-based sampling with natural language expert feedback to iteratively synthesize concise annotation guidelines, while requiring only the label definitions rather than an initial codebook~\citep{katz2024using, katz2026thematic}.

Our approach builds directly on the algorithm introduced by \citet{ranjit2025interventions}, with several important extensions. Specifically, \tool provides an interactive interface that enables users to edit the codebook at any point during development and extends the synthesizer to incorporate fine-grained feedback on predicted labels, supporting spans, reasoning, and natural language corrections. To support privacy-sensitive and high-impact applications, \tool can also be deployed with locally hosted open-source language models. Finally, we evaluate the system through two real-world case studies across two diverse application domains.

\section{Conclusion and Future Work}
\label{sec:conclusion}
\tool is a human-LM collaborative codebook development assistant that helps domain experts develop codebooks to uncover new insights from unstructured text data. 
By combining iterative language model annotation with targeted expert feedback, \tool reduces the expert effort involved in codebook development while maintaining downstream classification performance.
As future work, we plan to continue developing \tool as an open-source platform by incorporating feedback from community organizations and evaluating the system across a broader range of application domains.
Avenues for future work include examining how task parameters, such as sampling strategies and annotation budgets, influence downstream classification performance, as well as investigating alternative stopping criteria for human-in-the-loop codebook development, such as thematic saturation or adaptive stopping based on diminishing performance gains across iterations.
We envision \tool helping community organizations transform large collections of unstructured text into structured insights that support evidence-based decision making in high-impact domains.

\section{Limitations}

Our evaluation has several limitations. Although our case studies span distinct application domains, both formulate annotation as a single-target, three-class classification task. Future work should evaluate \tool on more complex settings, including larger or hierarchical label inventories, multi-label tasks, long documents, ambiguous or contested concepts, and workflows involving multiple experts.

Our evaluation focuses on the feasibility and downstream performance of \tool rather than systematically isolating the effects of individual design choices. Future work could evaluate additional baselines and examine the effects of prompting strategies, prompting ablations, sampling methods, annotation budgets, and stopping criteria. In particular, our current workflow does not explicitly measure thematic saturation (e.g., whether additional examples continue to surface new concepts or require substantive changes to the codebook). Future work could investigate thematic saturation or diminishing improvements across iterations as adaptive stopping criteria for codebook development. It would also be valuable to study qualitative differences between expert and \tool codebooks and how effectively each transfers across human and language model annotators.

\tool relies on a representative expert-annotated validation set to evaluate codebooks during development. Constructing such a set introduces additional expert effort, and its size and representativeness may influence both codebook selection and measured performance. Future work should investigate how smaller validation sets and alternative stopping criteria can reduce this requirement. The reductions in development time and expert effort observed in our case studies should be interpreted in the context of each domain and may vary with task complexity.

\section{Ethics Statement}

\subsection{Intended Usage of Language Model Annotations}

\tool is designed as a human-in-the-loop decision system to develop codebooks, rather than a fully automated annotation pipeline. Language model predictions are intended to assist domain experts in developing and refining annotation guidelines, not to replace expert judgment. The downstream scaled language model annotations are intended to augment expert judgment by surfacing data-driven insights, rather than autonomously inform high impact decision making. We do not recommend deploying the system in high-stakes settings without expert oversight, particularly when annotations may influence clinical, legal, or policy decisions.

\subsection{Data Privacy}
The intended application domains for \tool may involve sensitive, proprietary, or personally identifiable text. Accordingly, the system supports local deployment, allowing organizations to retain ownership of their data and comply with institutional privacy and data governance requirements. In our case studies involving homelessness case management data, all records were de-identified and sanitized to remove personally identifiable information prior to codebook development and evaluation. Data were used in accordance with an established data use agreement with the participating organizations. Additionally, all case studies were conducted using open-source language models, and all data were stored on secure academic servers.

\section*{Acknowledgments}
We thank Hannah Murray, Trent Kobielusz, Jeena Mahajan, Ezinne Nwankwo, My Friend’s Place, and Breaking Ground for their contributions to the co-design of \tool. We additionally thank My Friend’s Place for providing anonymized case note data and supporting the youth homelessness case study, and Hannah Murray, Trent Kobielusz, and Jeena Mahajan for their contributions to the wildlife trafficking case study. We thank the anonymous reviewers, area chairs, and meta-reviewers of the EMNLP 2026 System Demonstrations track for their thoughtful feedback.

\bibliography{custom}

\appendix
\appendix
\label{sec:appendix}

\section{Codesign with Community Organizations}
\label{app:codesign}
To inform the design of \tool, we conducted three participatory co-design sessions with practitioners from My Friend's Place and Breaking Ground (homelessness), and researchers specializing in AI for Wildlife Conservation. These sessions consisted of demonstrations of prototypes, and think-aloud walkthroughs. Feedback from these sessions directly informed the design of the system.

Participants from both My Friend's Place and Breaking Ground emphasized that many future users would be case managers and researchers with limited technical experience. One participant remarked, \textit{`breaking the workflow into smaller guided steps instead of showing everything at once would be helpful.'} Participants also expressed confusion around terminology such as \textit{ground truth}, \textit{multi-label}, and \textit{sampling}. This feedback resulted in guided onboarding pages before each stage of the workflow and simplified terminology throughout the interface.

Wildlife conservation researchers noted that \textit{`The label might actually be correct, but the reasoning is completely wrong,'} while another participant commented, \textit{`I'd want to see what part of the text it's actually using.'} Similarly, My Friend's Place participants requested clearer explanations of evaluation metrics before trusting model performance and a place to \textit{`write down rules as we notice them'} rather than reconstructing them afterwards. Breaking Ground participants focused on long-running annotation workflows, asking, \textit{`Can I stop halfway through and come back later?'} and suggesting configurable annotation budgets before beginning a task. This feedback resulted in supporting evidence spans, model reasoning, synthesized guidelines, integrated metric explanations, a live editable codebook, the Synthesizer module for consolidating expert feedback, resumable annotation sessions, persistent task libraries, progress indicators, configurable annotation budgets, and collapsible workflow stages. We will continue to iteratively refine the system through ongoing community participation, co-design, and user feedback as the platform is deployed across additional organizations and application domains.

\section{\tool Prompt Templates and Software Specification}
\label{app:prompts_software}
\begin{table}[ht]
\centering
\scriptsize

\fbox{%
\begin{minipage}{0.95\columnwidth}

\textbf{Generator Prompt}

\vspace{0.3em}

\ttfamily

You are an expert annotation model. Your job is to classify the input text for the following task.

Task name: \{task\_name\}

Task description: \{task\_description\}

Allowed labels: \{labels\}

Codebook guidelines: \{guidelines\}

Input text: \{text\}

Return exactly one JSON object:

\{"label":["one\_allowed\_label"],

"span\_text":"exact supporting span from the input text or No span text found",

"reason":"brief reason based only on the input text and codebook"\}

Rules:
Output valid JSON only., Use exactly the keys \texttt{label}, \texttt{span\_text}, and \texttt{reason}. Select exactly one label from the allowed labels., Do not invent labels. Quote the supporting span verbatim. If no supporting span exists, return \texttt{"No span text found"}. Keep the reasoning concise and evidence-based.

\end{minipage}
}

\caption{Template used by the Generator module. Variables enclosed in braces are populated dynamically at runtime.}
\label{tab:generator_prompt}

\end{table}
\begin{table}[ht]
\centering
\scriptsize

\fbox{%
\begin{minipage}{0.95\columnwidth}

\textbf{Synthesizer Prompt}

\vspace{0.3em}

\ttfamily

You are an expert instruction engineer. Given expert feedback on LM predictions, synthesize annotation guidelines that improve future performance.

Input:
sample\_text, ai\_labels, ai\_reasoning, ai\_span\_text,
ground\_truth\_labels, user\_feedback,
user\_label/span/reasoning\_feedback.

Generate rules that are: Generalizable and concise, Clear, instructional, and evidence-based, Resolve ambiguous label boundaries, Improve label, span, and reasoning quality
Return only:

\{"rules":["Rule 1...", "Rule 2...", ...]\}

\end{minipage}
}

\caption{Template used by the Synthesizer module. Expert feedback from each annotation batch is synthesized into generalized codebook rules.}
\label{tab:synthesizer_prompt}

\end{table}









\begin{table}[h]
\centering
\small
\begin{tabular}{ll}
\toprule
\textbf{Model} & \textbf{Specification} \\
\midrule
Meta-Llama-3-70B-Instruct &
70B parameters \\

Qwen2.5-32B-Instruct &
32B parameters \\

Qwen2.5-14B-Instruct &
14B parameters \\

\midrule
GPU & NVIDIA RTX A6000 \\
\# GPUs & 4 (inference) \\
\bottomrule
\end{tabular}
\caption{Language models evaluated in this work. Qwen1.5-32B-Instruct was used for codebook development, while all three models were used for evaluation.}
\label{tab:model_list}
\end{table}

The system follows a three-tier architecture: a React single-page application served as static assets behind a Caddy reverse proxy (which also terminates TLS), a Node.js/Express backend-for frontend that handles authentication, task orchestration, and data persistence to MongoDB, and a Python/FastAPI ML service responsible for embedding-based sampling (via FAISS over \texttt{all-mpnet-base-v2} embeddings), and LM inference. Table~\ref{tab:software-spec} summarizes the
  software versions. 
\begin{table}[ht]
\centering
\small
\begin{tabularx}{\columnwidth}{@{}p{2.6cm}X@{}}
\toprule
\textbf{Component} & \textbf{Version / Specification} \\
\midrule
Frontend & React 19.2, TypeScript 5.9, Vite 7.2, Node.js 20 (LTS) \\
Backend & Express 5.2, MongoDB Node.js Driver 7.0, JWT (\texttt{jsonwebtoken} 9.0), Argon2 \\
Python Stack & Python 3.11, FastAPI, uvicorn (ASGI), PyTorch (CPU build) \\
Embeddings & \texttt{all-mpnet-base-v2} \\
Vector Search & FAISS (CPU) \\
LLM Serving & Ollama (local/development), OpenRouter API (production) \\
\bottomrule
\end{tabularx}
\caption{Software components and versions used to implement \tool.}
\label{tab:software-spec}
\end{table}

\vspace{-0.5cm}

\section{Case Study Metadata}
\begin{figure}[h]
    \centering
    \begin{subfigure}[h]{0.49\columnwidth}
        \centering
        \includegraphics[width=\linewidth]{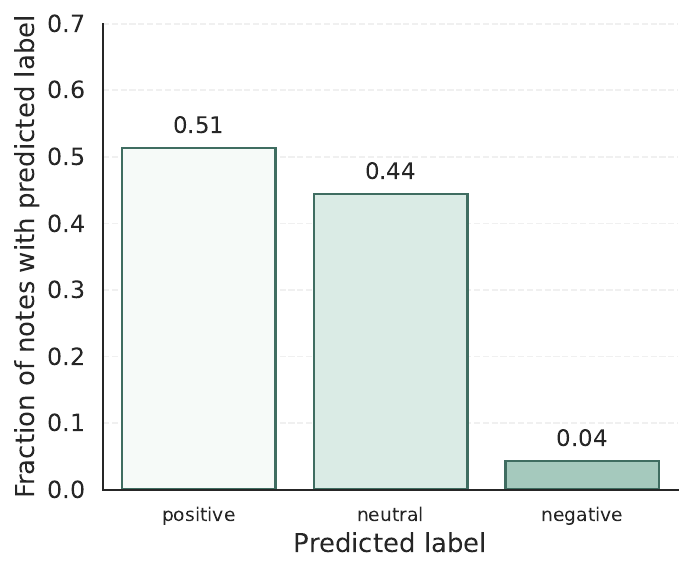}
        \caption{Wildlife Trafficking}
        \label{fig:wildlife_distribution}
    \end{subfigure}
    \hfill
    \begin{subfigure}[h]{0.49\columnwidth}
        \centering
        \includegraphics[width=\linewidth]{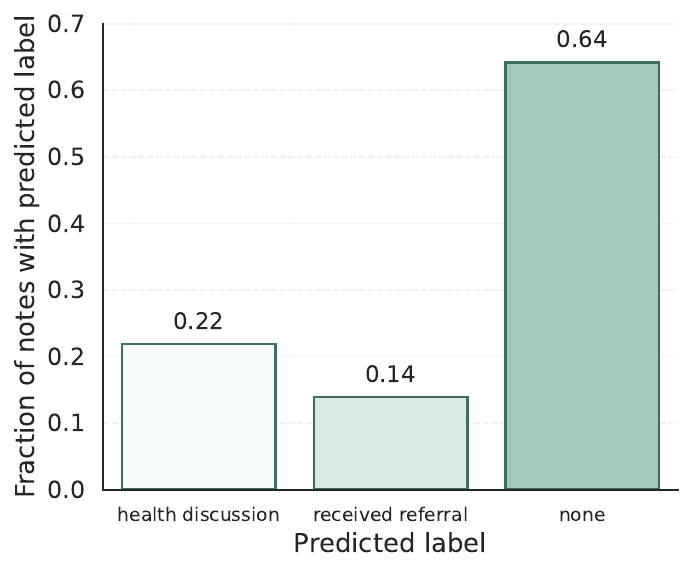}
        \caption{My Friend's Place}
        \label{fig:mfp_distribution}
    \end{subfigure}
    \caption{Distribution of Llama-3-70B predicted sentiment labels of 237K social media posts on wildlife trafficking of pangolins (left) and Qwen2.5-32B referral related labels of 6K case management notes (right). }
    \label{fig:predicted_label_distributions}
\end{figure}

We include the final distribution of scaled annotations on $\mathcal{D}_{\text{all}}$ in \autoref{fig:predicted_label_distributions}, \tool time statistics and user ratings on the usefulness of the explanations and span in \autoref{tab:annotation_quality}, dataset statistics for each of our case studies in \autoref{tab:mfp_dataset} and \autoref{tab:wildlife_dataset}, and a subset of guidelines comparing the expert codebook and \tool codebook in \autoref{tab:mfp_codebook}. 
\label{app:casestudy}

\begin{table}[ht]
\centering
\small
\begin{tabular}{lrrr}
\toprule
\textbf{Statistic} & $\mathbf{D^{30}_{\text{guide}}}$ & $\mathbf{D_{\text{val}}}$ & $\mathbf{D_{\text{all}}}$ \\
\midrule
Total Examples & 150 & 585 & 237,977 \\
\midrule
Positive & 36 & 128 & 122,160 \\
Negative & 44 & 251 & 10,196 \\
Neutral & 70 & 206 & 105,621 \\
\bottomrule
\end{tabular}
\caption{Dataset statistics and label distributions for the wildlife trafficking sentiment classification task. }
\label{tab:wildlife_dataset}
\end{table}

\begin{table}[t]
\centering
\small
\begin{tabular}{lrrr}
\toprule
\textbf{Statistic} & $\mathbf{D^{30}_{\text{guide}}}$ & $\mathbf{D_{\text{val}}}$ & $\mathbf{D_{\text{all}}}$ \\
\midrule
Total Examples & 150 & 593 & 6,725 \\
\midrule
Not Relevant & 114 & 416 & 4,315 \\
Health Discussion & 12 & 87 & 1,474 \\
Received Referral & 24 & 90 & 936 \\
\bottomrule
\end{tabular}
\caption{Dataset statistics and label distributions for the My Friend's Place health referral case study. }
\label{tab:mfp_dataset}
\end{table}
\begin{table}[H]
\centering
\small
\begin{tabular}{lcc}
\toprule
\textbf{Metric} & \textbf{MFP} & \textbf{Wildlife} \\
\midrule
Total Time (min) & 114.0 & 55.3 \\
LM Explanation Useful (\%) & 73.2 & 54.0 \\
LM Span Useful (\%) & 71.8 & 96.7 \\
\bottomrule
\end{tabular}
\caption{Annotation efficiency and qualitative feedback collected with \tool.}
\label{tab:annotation_quality}
\vspace{-1.5em}
\end{table}

\begin{table}[H]
\centering
\scriptsize

\fbox{%
\begin{minipage}{0.95\columnwidth}

\textbf{Expert Codebook $\mathcal{G}_{\text{expert}}$}

\vspace{0.3em}

{\ttfamily

\textbf{Health Discussion}

-- Mental health only does not count. Discussion could be physical and mental health related.

-- Even if only one sentence about physical health out of a long note, it counts.

-- Case notes exclusively limited to between providers are ``none'' because the young person is not involved in the discussion.
\ldots

\medskip

\textbf{Referral}

-- If assistance is recommended/offered, then it counts as a referral.

-- Even if young person declines, still counts as referral (e.g., ``declined medical support'').

-- Offering transportation, care coordination involving young person (e.g., creating appointment, checking insurance coverage, logistics planning/support).
\ldots

\medskip

\textbf{Not Relevant}

-- Medication should be coded as ``none'' because do not know if it's for physical or mental health, unless specified or based on context clues.

-- Reference to hospital may not be clear if it's for physical or mental health. This results in a ``none'' code.
\ldots

}

\end{minipage}
}

\vspace{0.8em}

\fbox{%
\begin{minipage}{0.95\columnwidth}

\textbf{\tool\ Codebook ($\mathcal{G}^{23}_{\tool}$)}

\vspace{0.3em}

{\ttfamily

\textbf{Health Discussion}

-- Health discussions must involve clear evidence of an individual's health status and its potential impact, not just general observations; if the text lacks explicit mention of treatment or referrals, reconsider the label.

-- When the text primarily discusses personal conversations or administrative tasks without clear healthcare intervention, label as ``not relevant'' instead of ``health discussion''.
\ldots

\medskip

\textbf{Referral}

-- Consider the context and actions taken; if a case manager coordinates or arranges a meeting or appointment for the individual with healthcare services, label it as ``referral'', even if no specific medical care is provided.

-- Include explicit mention of an individual's interaction with healthcare services to justify a ``referral'' label. If the individual declines medical care, still code the event as a ``referral''.
\ldots

\medskip

\textbf{Not Relevant}

-- If the case note focuses on a discussion between agencies without addressing specific treatment or referrals for health concerns, label as ``not relevant''.

-- Do not label as ``referral'' when discussing coordination between agencies without direct involvement in securing appointments.
\ldots

}

\end{minipage}
}

\caption{Comparison of representative expert and \tool annotation guidelines for the My Friend's Place case study.}
\label{tab:mfp_codebook}
\end{table}

\end{document}